\documentclass[preprint,12pt]{elsarticle}

\usepackage{amsmath, amssymb}
\usepackage{booktabs}      
\usepackage{colortbl}     
\usepackage{multirow}      
\usepackage{array}         
\usepackage[table,xcdraw]{xcolor} 
\usepackage{float}
\usepackage{pifont}
\newcommand{\cmark}{\ding{51}} 
\newcommand{\xmark}{\ding{55}}
\usepackage{algorithm}
\usepackage{algpseudocode}

\usepackage[normalem]{ulem}

\usepackage[colorlinks=true, linkcolor=blue, citecolor=blue, urlcolor=blue]{hyperref}

\journal{Information Fusion}

\begin{document}

\begin{frontmatter}

\title{EH-Benchmark: Ophthalmic Hallucination Benchmark and Agent-Driven Top-Down Traceable Reasoning Workflow}

\author[1,2]{Xiaoyu Pan}
\author[1]{Yang Bai\corref{cor1}}
\author[2]{Ke Zou}
\author[1]{Yang Zhou}
\author[1]{Jun Zhou}

\author[1]{Huazhu Fu}
\author[2,3]{Yih-Chung Tham}

\author[1]{Yong Liu}

\address[1]{Institute of High Performance Computing, Agency for Science, Technology and Research (A*STAR), 1 Fusionopolis Way, Singapore, 138632}

\address[2]{Centre for Innovation and Precision Eye Health; and Department of Ophthalmology, NUHS Tower Block, Level 7, 1E Kent Ridge Road, Singapore, 119228}

\address[3]{Singapore Eye Research Institute, Singapore National Eye Centre, 20 College Road, Singapore, 169856}

\cortext[cor1]{Corresponding author. Email: Bai\_Yang@ihpc.a-star.edu.sg}

\begin{abstract}
Medical Large Language Models (MLLMs) play a crucial role in ophthalmic diagnosis, holding significant potential to address vision-threatening diseases. However, their accuracy is constrained by hallucinations stemming from limited ophthalmic knowledge, insufficient visual localization and reasoning capabilities, and a scarcity of multimodal ophthalmic data, which collectively impede precise lesion detection and disease diagnosis. Furthermore, existing medical benchmarks fail to effectively evaluate various types of hallucinations or provide actionable solutions to mitigate them. To address the above challenges, we introduce EH-Benchmark, a novel ophthalmology benchmark designed to evaluate hallucinations in MLLMs. We categorize MLLMs' hallucinations based on specific tasks and error types into two primary classes: Visual Understanding and Logical Composition, each comprising multiple subclasses. Given that MLLMs predominantly rely on language-based reasoning rather than visual processing, we propose an agent-centric, three-phase framework, including the Knowledge-Level Retrieval stage, the Task-Level Case Studies stage, and the Result-Level Validation stage. Experimental results show that our multi-agent framework significantly mitigates both types of hallucinations, enhancing accuracy, interpretability, and reliability. Our project is available at https://github.com/ppxy1/EH-Benchmark.

\end{abstract}



\end{frontmatter}

\section{Introduction}
According to the World Health Organization (WHO), eye conditions have a major global impact, affecting at least 2.2 billion people with vision problems. Low- and middle-income areas face a greater burden, with rates four times higher than in high-income regions. Since most conditions show no early signs, they often remain unnoticed. Additionally, many people lack knowledge about eye health and skip regular checkups \cite{foot2017surveillance}. 

To tackle diagnostic challenges, researchers have developed computer-aided diagnosis (CAD) methods, utilizing automated detection and recognition to reduce clinicians' workload and provide efficient patient services. Recently, CAD has been widely applied in ophthalmic tasks, such as lesion detection and disease progression prediction. Despite high accuracy, current CAD systems still require physician oversight, which limits their broader adoption.

In recent years, Medical Large Language Models (MLLMs) have gained attention for their ability to process diverse medical data, including imaging modalities (e.g., MRI, CT scans), time-series signals (e.g., outputs from electroencephalogram), audio recordings (e.g., patient interviews), and clinical documentation \cite{alsaad2024multimodal,thirunavukarasu2023large}. These models exhibit strong performance in addressing complex medical queries based on multimodal inputs \cite{lin2025self,luo2023chatkbqa}, generating structured medical reports from raw clinical data \cite{wang2025llm,li2025towards}, and recommending personalized treatment plans grounded in patients’ genetic profiles and clinical histories \cite{galitsky2024llm,benary2023leveraging}. Despite their promising capabilities, MLLMs often produce factually incorrect yet seemingly plausible content, a phenomenon known as hallucination. This typically results from a lack of deep domain knowledge, weak reasoning, or misinterpretation of context \cite{xu2025knowledge}. In addition, their decision-making processes lack transparency, making it difficult to trace or verify their outputs. These limitations are particularly concerning in the medical field, where accuracy and interpretability are essential.

Although MLLMs have advanced quickly, solid ophthalmology benchmarks remain scarce. Most studies focus on text-based tasks, such as knowledge quizzes, open-ended questions, and multiple-choice tests, while few have investigated or evaluated the hallucinations produced by LLMs in ophthalmic settings \cite{da2025integrated, li2025eyecaregpt}. Existing benchmarks usually cover only one data type or a single task, so they fail to capture the mix of images, text, and other signals used in real clinics. As Table~\ref{Table1} shows, this lack of broad, multimodal benchmarks makes it hard to compare methods fairly and slows progress. Creating an ophthalmology benchmark that combines varied data types and task formats is therefore an urgent research need.

In this paper, we introduce a novel ophthalmic benchmark comprising \textbf{13} datasets and \textbf{3} modalities. It addresses two major categories of hallucination—Visual Understanding Hallucination and Logical Composition Hallucination—across three levels of clinical reasoning: instance, pathological, and decision-making, totaling \textbf{27K} questions. Visual Understanding Hallucination is further divided into five error types: Numerical, Categorical, Positional, Diagnosis-Type, and Stage-Level Errors. Logical Composition Hallucination consists of \textbf{806} questions derived from over \textbf{200} published case reports, specifically curated to evaluate the reasoning capabilities of large language models (LLMs). The proposed framework is extensible and can be directly applied to construct new benchmark datasets.

To address the issue of hallucinations in LLMs, researchers have begun exploring the use of Agents. However, most existing approaches rely on general frameworks that simulate doctor-patient interactions to optimize diagnostic reasoning. These methods are limited by significant deficiencies in contextual understanding and a lack of interpretability, which obstruct the seamless integration of Agents with LLMs. Furthermore, since many LLMs predominantly utilize pattern recognition and correlations between symptoms—rather than reconstructive reasoning based on clinical guidelines—they encounter substantial difficulties in managing complex cases or diagnosing rare diseases. To overcome these challenges, we have developed a multi-agent framework specifically tailored for ophthalmic tasks. This framework is structured around three key stages: Knowledge-Level Retrieval, Task-Level Case Studies, and Result-Level Validation. In this initial phase, external data sources are leveraged to pair queries with the contextual backgrounds of relevant cases, enabling the retrieval of clinical guidelines pertinent to ophthalmology. During the middle phase, a Decision Agent is employed to select and sequence professional tools, constructing a diagnostic workflow that ensures logical consistency throughout the process. In the final phase, an Evaluation Agent conducts a thorough assessment of the tool outputs across three critical dimensions: correctness, completeness, and adherence to the predefined workflow. The system addresses identified deficiencies by selectively retrying specific tool agents, establishing an iterative self-correction loop. Through this structured process, the diagnostic system evolves from an opaque, black-box model into a clinically transparent, self-correcting, and trustworthy AI assistant. The main contributions are as follows:

\begin{itemize}
\item We introduce EH-Benchmark, a new ophthalmology benchmark with over \textbf{27K} ophthalmic questions, designed to evaluate Visual Understanding and Logical Composition hallucinations induced by LLMs.

\item To address hallucinations in ophthalmic tasks, we propose a multi-agent framework comprising three key stages: Knowledge-Level Retrieval, Task-Level Case Studies, and Result-Level Validation.

\item In the first stage, external ophthalmic knowledge sources are queried to improve retrieval relevance. During the second stage, a Decision Agent selects tools and their invocation order, cyclically building the diagnostic workflow. In the final stage, an Evaluation Agent ensures the workflow is accurate, complete, and consistent, enabling a transparent and reliable AI assistant.

\item Experiments show that our framework achieves state-of-the-art performance on the EH-Benchmark, surpassing other LLMs, thereby highlighting its potential for visual understanding and reasoning.
\end{itemize}

\section{Literature Review}
\subsection{LLM-based AI Agents}
LLM-based agents are evolving into autonomous, multimodal systems with tool-leveraging capabilities, yet they face persistent challenges in memory, deployment, and safety \cite{xi2025rise}. Architectures like Talebirad et al.'s collaborative framework \cite{talebirad2023multi} assign specialized roles to agents in AGI applications, which improves complex task handling. However, dynamic team coordination within such frameworks often encounters scalability issues. Innovations such as DyLAN's temporal networks \cite{liu2024dynamic} and Guo et al.'s leadership hierarchies \cite{guo2024embodied} enhance adaptability, boosting code generation and decision-making accuracy. However, task–agent alignment remains problematic and frequently causes conflicts. Meanwhile, tool integration advances, with TOOLMAKER's code repository conversion \cite{wolflein2025llm} and Wang et al.'s CodeAct framework \cite{wang2024executable} utilizing Python interoperability to achieve a 20\% performance increase in complex operations. However, domain-specific robustness lags, particularly in healthcare, where hallucination, fragmented tool ecosystems, and weak reasoning validation expose vulnerabilities, underscoring the need for verifiable, domain-tailored frameworks to ensure reliable, safe deployment in critical settings.

\begin{table}[H]
\centering
\fontsize{7}{10}\selectfont
\caption{Comparison of existing general-domain and medical-domain benchmarks in evaluating LLM performance across modalities, image types, and hallucination dimensions.}
\label{Table1}
\vspace{10pt} 
\begin{tabular}{l|cc|cc|c|cccccc}
\toprule
\textbf{Benchmark} & \multicolumn{2}{c|}{\textbf{Modalities}} & \multicolumn{2}{c|}{\textbf{Image Type}} & \textbf{\#Num} & \multicolumn{6}{c}{\textbf{Hallucination}} \\
\cmidrule(lr){2-3} \cmidrule(lr){4-5} \cmidrule(lr){7-12}
& \textbf{Images} & \textbf{Texts} & \textbf{CFP} & \textbf{OCT} & \centering\arraybackslash \textbf{} & \textbf{Num} & \textbf{Cat} & \textbf{Pos} & \textbf{Diag} & \textbf{Sta} & \textbf{Clin} \\
\midrule
\multicolumn{12}{c}{\textbf{General-Domain Benchmarks}} \\
\midrule
MMMU \cite{yue2024mmmu} & \cmark & \cmark & \xmark & \xmark & \textbf{11.5K} & \xmark & \xmark & \xmark & \xmark & \xmark & \xmark \\
MathVista \cite{lu2023mathvista} & \cmark & \cmark & \xmark & \xmark & \textbf{6K} & \xmark & \xmark & \xmark & \xmark & \xmark & \xmark \\
UNK-VQA \cite{guo2024unk} & \cmark & \cmark & \xmark & \xmark & \textbf{10K} & \xmark & \xmark & \xmark & \xmark & \xmark & \xmark \\
\midrule
\multicolumn{12}{c}{\textbf{Medical-Domain Benchmarks}} \\
\midrule
LLMEval-Med \cite{zhang2025llmeval} & \xmark & \cmark & \xmark & \xmark & \textbf{3K} & \xmark & \xmark & \xmark & \xmark & \xmark & \xmark \\
MedHEval \cite{chang2025medheval} & \cmark & \cmark & \xmark & \xmark & \textbf{15K} & \xmark & \cmark & \cmark & \cmark & \xmark & \xmark \\
HEAL-MedVQA \cite{nguyen2025localizing} & \cmark & \cmark & \xmark & \xmark & \textbf{67K} & \xmark & \cmark & \cmark & \cmark & \xmark & \xmark \\
FunBench \cite{wei2025funbench} & \cmark & \cmark & \cmark & \cmark & \textbf{91K} & \cmark & \cmark & \cmark & \cmark & \xmark & \xmark \\
LMOD \cite{qin2024lmod} & \cmark & \cmark & \cmark & \cmark & \textbf{21K} & \xmark & \cmark & \xmark & \cmark & \xmark & \xmark \\
\textbf{EH-Benchmark} & \cmark & \cmark & \cmark & \cmark & \textbf{27K} & \cmark & \cmark & \cmark & \cmark & \cmark & \cmark \\
\bottomrule
\end{tabular}
\end{table}

\subsection{Medical Large Language Models}
Large language models are revolutionizing healthcare by enhancing diagnosis, treatment planning, and clinical decision-making \cite{he2025survey,lin2024has}. Recent advancements in MLLMs are driving a shift from general medical tasks to specialized, task-specific challenges. Sam-u \cite{deng2023sam} and Medsam-u \cite{zhou2024medsam} develop robust prompts to address the challenges posed by different prompt types and their spatial sensitivity. BiomedGPT \cite{zhang2024generalist} and MedGemma \cite{medgemma-hf} introduce lightweight, open-source vision-language models for biomedical tasks and medical image understanding, achieving state-of-the-art performance and enhancing diagnostic accuracy. For biomedical image analysis, LLaVA-Med \cite{li2023llava} delivers superior multimodal conversational capabilities with low training costs, while MENDER \cite{lin2025cross} excels in differential-aware medical VQA through cross-modal knowledge diffusion and multi-scale medical knowledge fusion. In medical consultation, HuatuoGPT \cite{chen2024huatuogptvisioninjectingmedicalvisual} leverages distilled ChatGPT data and real-world clinical data with reinforcement learning, achieving leading performance among open-source LLMs. However, the continued lack of high-quality datasets remains a major barrier to further progress and widespread application of MLLMs in precision medicine.

\subsection{Hallucination in Medical Large Language Models}
The emergence of MLLMs has significantly advanced medical applications, particularly in medical image interpretation and diagnostic visual question answering. However, their clinical deployment is hindered by persistent hallucinations—generating pathologically or anatomically incorrect content not substantiated by input data \cite{yan2024evaluating}. While frameworks such as MedVH \cite{gu2024medvh} and Med-HallMark \cite{chen2024detecting} have pioneered hallucination detection, their focus remains predominantly on visual modality errors, overlooking critical hallucinations stemming from knowledge deficiencies and contextual misalignments—factors essential for accurate clinical decision-making. Current approaches reveal two primary limitations: (1) existing taxonomies fail to adequately capture the epistemological roots of hallucinations, particularly in distinguishing modality-specific artifacts (e.g., misinterpreting radiological images) from domain knowledge inadequacies (e.g., misapplied anatomical ontologies); (2) emerging benchmarks like MedHallBench \cite{zuo2024medhallbench} and MedHEval \cite{chang2025medheval} are limited by their narrow emphasis on visual artifacts, insufficient granularity in error classification, and lack of clinically valid metrics. This gap underscores the urgent need for multidimensional evaluation frameworks that systematically address hallucinations across visual perception, biomedical knowledge representation, and clinical context integration, a prerequisite for developing clinically reliable MLLMs.

\section{EH-Benchmark}
\label{eh}

Based on the specific tasks and error types observed in the ophthalmology domain, we classify hallucinations into \textbf{Visual Understanding Hallucination (A1)} and \textbf{Logical Composition Hallucination (A2)}. An overview of the proposed EH‑Benchmark is provided in Figure~\ref{fig1}. A1 primarily involves visual perception errors, whereas A2 reflects reasoning errors in multimodal knowledge integration, representing a form of compositional hallucination. In the following sections, we provide a detailed analysis of the underlying causes of these two types of ophthalmic hallucinations. All questions are multiple-choice, and model predictions are evaluated using regular-expression matching. Task details are shown in Table~\ref{Table2}. To illustrate representative hallucinations produced by large language models across different tasks, we select several prototypical questions, decompose their model‑generated answers, and contrast them with the ground‑truth responses (see Figure~\ref{fig22}).

\begin{figure}[H]
  \centering
  \includegraphics[width=0.9\linewidth]{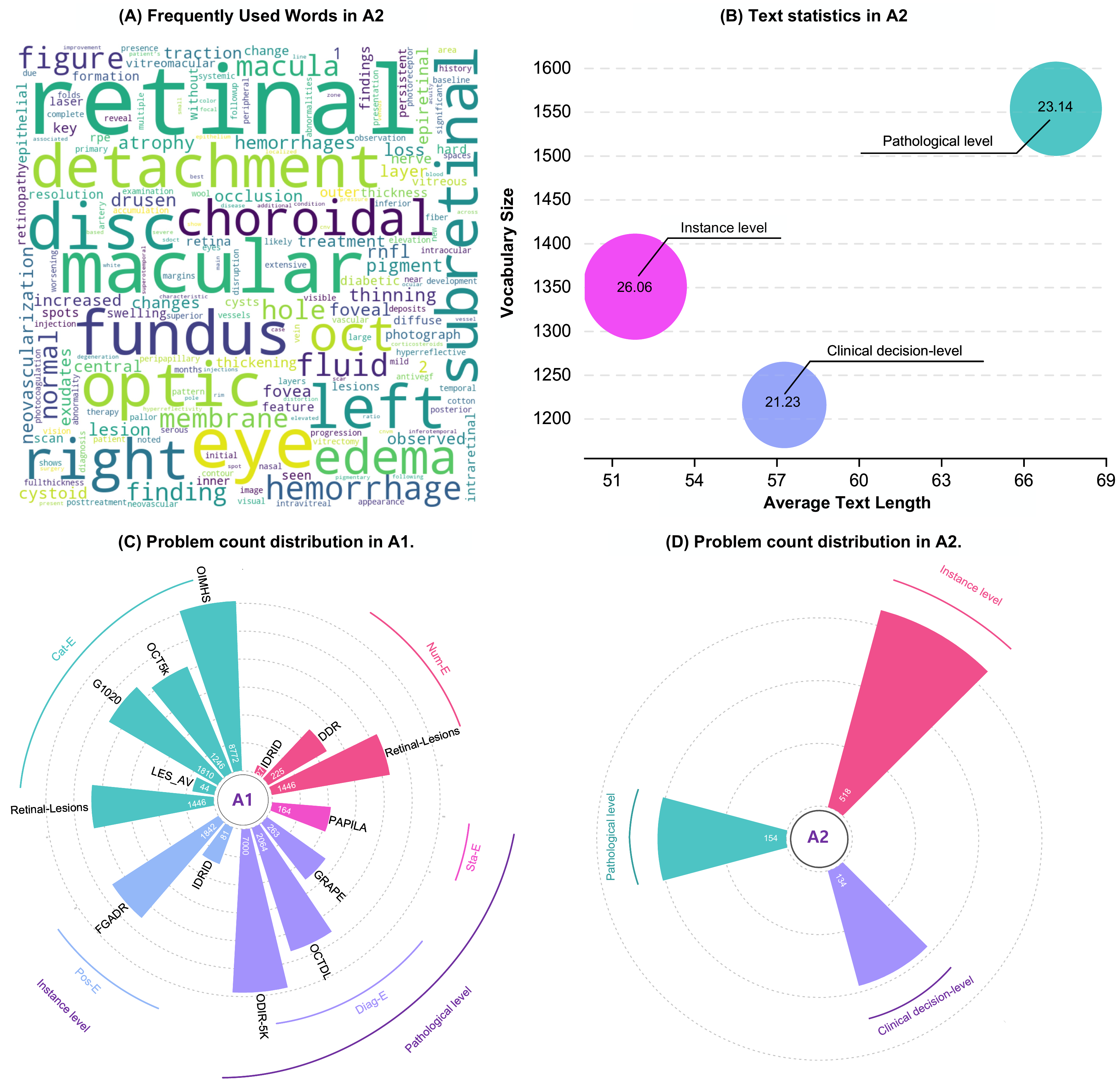}
  \caption{Statistical overview of our proposed EH-Benchmark. (A) Word clouds of questions and options in the A2 task, illustrating the diversity of ophthalmology-related vocabulary. (B) A comprehensive evaluation of the average text length, vocabulary size, and vocabulary diversity across the three types of ophthalmic hallucinations in the A2 task. (C) Statistics on the two types of hallucinations in the A1 task, along with the number of questions in each subcategory. (D) Number of questions for the three types of hallucinations in the A2 task.}
  \label{fig2}
\end{figure}

\subsection{A1: Visual Understanding Hallucination}
\label{a1}

The Visual Understanding Hallucination arises when the model generates explanations involving visual features that are not present or lack empirical grounding. This typically occurs as the model attempts to infer fine-grained pathological details—such as claiming the presence of retinal hemorrhages or cotton-wool spots in a Color Fundus Photography (CFP) — despite no corresponding visual evidence. Such hallucinations underscore the model’s limitations in grounding visual predictions in actual input data, revealing a gap between learned visual-textual associations and factual visual recognition.

To evaluate the severity of Visual Understanding Hallucinations across different LLMs, we collected approximately 26k questions from 12 ophthalmic visual datasets, namely: Retinal-Lesions \cite{wei2021learn}, DDR \cite{li2019diagnostic}, IDRID 
 \cite{porwal2018indian}, OIMHS \cite{ye2023oimhs}, OCT5k \cite{arikan2025oct5k}, G1020 \cite{bajwa2020g1020}, LES\_AV \cite{orlando2018towards}, FGADR \cite{zhou2020benchmark}, ODIR-5K \cite{odir2019}, OCTDL\cite{kulyabin2024octdl}, GRAPE \cite{huang2023grape}, and PAPILA 
 \cite{kovalyk2022papila}. The composition of questions in the A1 task is shown in Figure~\ref{fig2}(C).

Unlike existing ophthalmology benchmarks, we focus on five visual tasks across two levels of granularity, instance level and pathological level, to evaluate the phenomenon of Visual Understanding Hallucination. 

\textbf{Instance-level hallucination} refers to the model's generation of hallucinated outputs when recognizing or describing specific instances in an image, highlighting the limitations of LLMs in understanding fine-grained visual details.

\begin{enumerate}
    \setcounter{enumi}{0}
    \item \textbf{\textit{Numerical Error}}: The model exhibits visual hallucinations when counting specific features in medical images (such as the number of hemorrhages), and miscounting these features may lead doctors to overestimate or underestimate the severity of the condition, thereby affecting treatment decisions.
\end{enumerate}


\begin{enumerate}
    \setcounter{enumi}{1}
    \item \textbf{\textit{Categorical Error}}: The model shows hallucinations when identifying specific types of lesions\slash organs in medical images, being susceptible to interference from morphological similarities among lesions or image noise. Misidentification of lesion locations may lead to misdiagnosis, such as treating retinal lesions as choroidal ones.
\end{enumerate}



\begin{enumerate}
    \setcounter{enumi}{2}
    \item \textbf{\textit{Position Error}}: The model tends to generate hallucinations during the localization of lesions or specific regions in medical images, often caused by ambiguous spatial reference points or limited spatial reasoning capabilities. Inaccurate localization may impair the precision of surgical planning or targeted interventions.
\end{enumerate}


\textbf{Pathological level hallucination} occur when LLMs generate false content during the diagnosis or synthesis of an overall pathological condition. These hallucinations are influenced not only by the learning capacity of LLMs but also by the type and severity of the disease.

\begin{enumerate}
    \setcounter{enumi}{0}
    \item \textbf{\textit{Diagnosis-Type Error}}: The model may generate hallucinations when diagnosing the type of lesion presented in medical images, primarily due to insufficient training data or imbalanced datasets. Incorrect disease classification may result in unnecessary treatment or delays in appropriate medical intervention.
\end{enumerate}


\begin{enumerate}
    \setcounter{enumi}{1}
    \item \textbf{\textit{Stage-Level Error}}: The model may produce hallucinations when assessing disease severity or grading, caused by the complexity of ophthalmic grading standards or the model’s insensitivity to subtle lesion progression. Grading errors may lead to mismatched treatment intensity, such as overtreatment of patients with mild glaucoma.
\end{enumerate}


\subsection{A2: Logical Composition Hallucination}
The Logical Composition Hallucination focuses on errors in logical structure and reasoning during text generation by LLMs. This type of hallucination serves to evaluate the model’s ability to integrate complex information and may be related to its depth of reasoning and contextual understanding. In multi-step ophthalmic reasoning tasks, for example, a model may correctly perform the initial steps but later produce conclusions that contradict earlier inferences during deeper analysis.

Most existing ophthalmic visual-question-answering (VQA) datasets are directly derived from visual data (see Subsection \ref{a1}) and primarily focus on descriptive tasks, often lacking contextual information such as patient history or symptoms. This limitation makes them inadequate for supporting complex clinical reasoning. To better evaluate logical compositional hallucinations—i.e., logical errors that occur during reasoning—in LLMs, we collect relevant questions from over 200 published case reports. Case reports provide detailed clinical scenarios, including patient history, symptoms, and diagnoses, offering a more realistic and context-rich foundation for evaluation. Notably, all answer sentences in our benchmark are directly traceable to the original case report text. Each question is designed to require integration of visual cues and textual information, effectively testing the model’s logical compositional abilities. Two medically trained reviewers score each question and its answer choices based on reliability, completeness, and the type of hallucination present. Any low-quality questions are removed. After this process, we build a final dataset with 806 carefully selected questions. Figure~\ref{fig2}(D) shows the composition of questions in the A1 task. From Figure~\ref{fig2}(A) and (B), we observe that the A2 task exhibits high diversity and rich ophthalmic vocabulary.

\textbf{Instance-Level Hallucination}: Similar to A1, A2 also emphasizes hallucination at the level of specific features or instances within an image. In contrast, A2 focuses on the coherence of the reasoning process at the instance level, specifically addressing how reasonable conclusions can be derived based on image evidence. Some questions necessitate the comparison of multiple images from the same phase to enhance the robustness of the inference.


\textbf{Pathological-Level Hallucination}: In contrast to A1, which prioritizes the pathological accuracy of image content (i.e., the correct identification of pathological features), A2 builds upon A1 by emphasizing logical reasoning based on these features rather than mere feature recognition. Initially, LLMs must identify pathological features and subsequently correlate them with specific pathological conditions to produce a logically coherent pathological description. The challenge lies in the potential for the model to accurately identify certain features but falter in logically associating them with the definitions of pathological conditions. Furthermore, errors in inferring pathological conditions may arise due to insufficient medical knowledge or disruptions in the reasoning chain.

\textbf{Clinical Decision-Level Hallucination}: Distinct from the preceding two hallucination categories, this type emerges due to inaccuracies in LLMs' assessment of therapeutic outcomes. It necessitates that LLMs grasp pathological progression and clinical protocols to facilitate dynamic inference grounded in temporal data sequences.



\begin{figure}[H]
  \centering
  \includegraphics[width=0.9\linewidth]{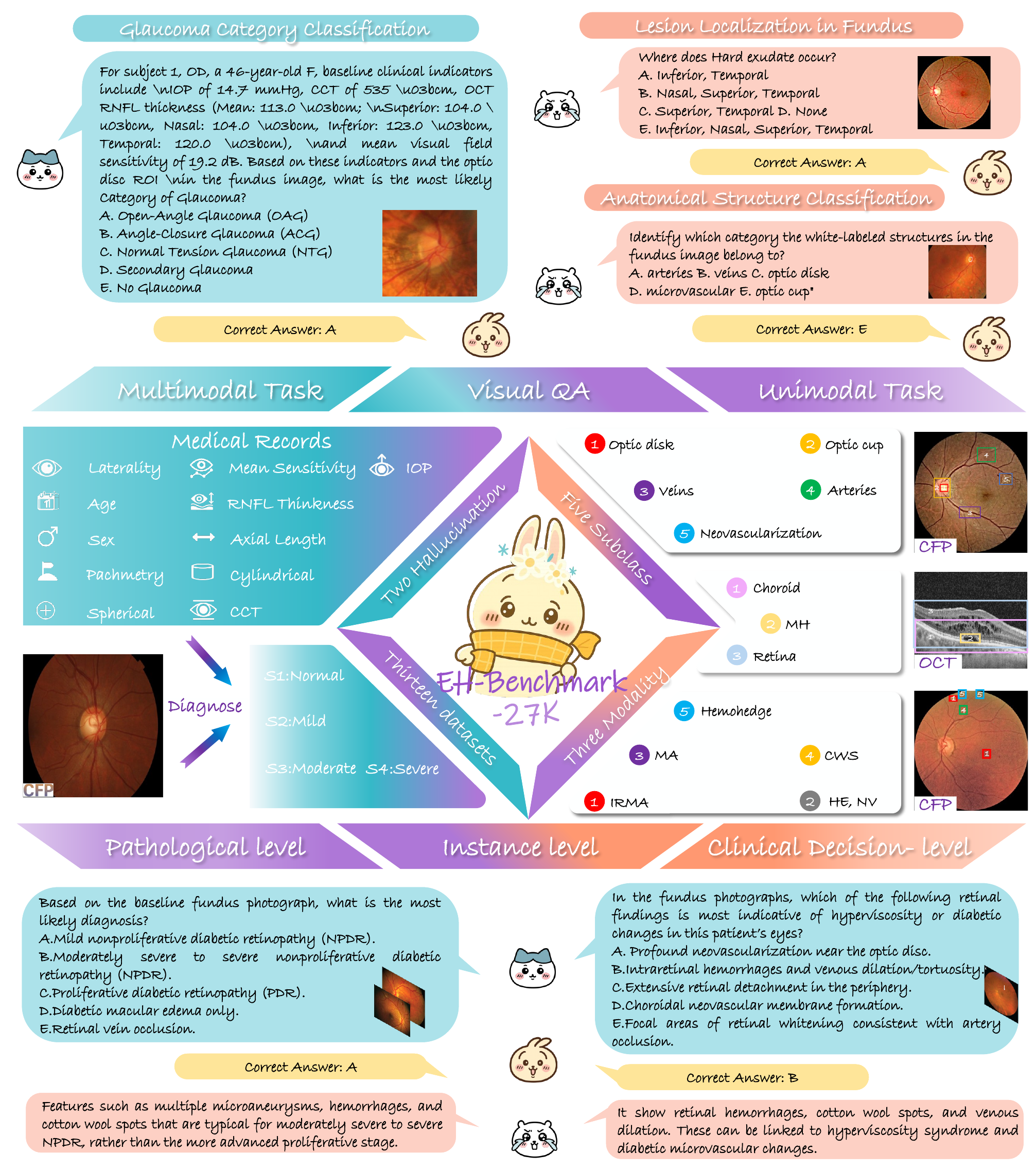}
  \caption{Overview of the EH-Benchmark, which aggregates data from 3 modalities and 13 datasets for ophthalmic vision QA tasks.}
  \label{fig1}
\end{figure}

\begin{table}[htbp]
  \centering
  {\fontsize{8}{10}\selectfont
   \setlength{\tabcolsep}{4.5pt}
   \renewcommand{\arraystretch}{1.3}
  \caption{Introduction to ophthalmic tasks.  
  The lesion-analysis task is based on three datasets: \textit{IDRID}, \textit{DDR}, and \textit{Retinal-Lesions}.}
  \label{Table2}
  \vspace{6pt}
  \begin{tabular}{llll}
    \hline
    Dataset & Modality & \#Num & Task description \\
    \hline
    PAPILA            & CFP, Text & 164   & Medical records + images, glaucoma severity grading \\
    Retinal-Lesions   & CFP & 1,446 & DR lesion-progression grading \\
    \rowcolor[HTML]{E6E0EC}
    FGADR             & CFP       & 1,842 & Identification of fundus lesions (HE, SE, etc) \\
    \rowcolor[HTML]{E6E0EC}
    IDRID             & CFP       & 81    & Localization of fundus lesions (MAs, Hemorrhages, FE, SE) \\
    Lesion analysis               & CFP       & 1,698 & Counting of fundus lesions (SE, MAs, Hemorrhages, etc) \\
    G1020             & CFP       & 1,810 & Optic cup / disc identification \\
    OCT5k             & OCT       & 1,246 & OCT lesion localization (PRLD, Fluid accumulation, etc) \\
    LES-AV            & CFP       & 44    & Artery / vein recognition \\
    OIMHS             & OCT       & 8,772 & Choroid, retina and macular-hole recognition \\
    \rowcolor[HTML]{DBEEF3}
    GRAPE             & CFP, Text       & 263   & Medical records + images, glaucoma-type identification \\
    \rowcolor[HTML]{DBEEF3}
    OCTDL             & OCT       & 2,064 & Diagnosis of seven common OCT diseases (AMD, DME, etc.) \\
    \rowcolor[HTML]{DBEEF3}
    ODIR-5K           & CFP       & 7,000 & Diagnosis of fundus diseases (atrophy, pigment changes, etc.) \\
    Case Report       & CFP, OCT, Text & 806 & Comprehensive ophthalmological questions from case reports \\
    \hline
  \end{tabular}}
\end{table}

\section{Methods}
To address the challenge of mitigating various types of ophthalmic hallucinations, we propose a novel multi-agent framework structured into three distinct stages: Knowledge-Level Retrieval, Task-Level Case Studies, and Result-Level Validation. The comprehensive workflow is illustrated in Figure~\ref{fig3}. In the Knowledge-Level Retrieval stage, the RAG Agent extracts pertinent case backgrounds from an ophthalmic database, supplying a wealth of domain-specific knowledge for subsequent analyses. During the Task-Level Case Studies stage, the Decision Agent leverages the query and the case background information retrieved in the previous stage to optimally select and sequence appropriate tools, ensuring an effective analytical process. The invocation results of each tool are returned to facilitate subsequent evaluation. In the Result-Level Validation stage, the Evaluation Agent assesses the correctness, completeness, and adherence to the planned workflow, initiating retries as necessary to ensure diagnostic accuracy. As illustrated in Algorithm 1, the input-feedback loop keeps iterating until the system either produces a final response or determines that additional information is needed from the patient or doctor. In the following sections, we provide a detailed introduction to all the agents and tools involved, accompanied by an in-depth analysis of the three-stage workflow.

\begin{figure}[htbp]
  \centering
  \includegraphics[width=\linewidth]{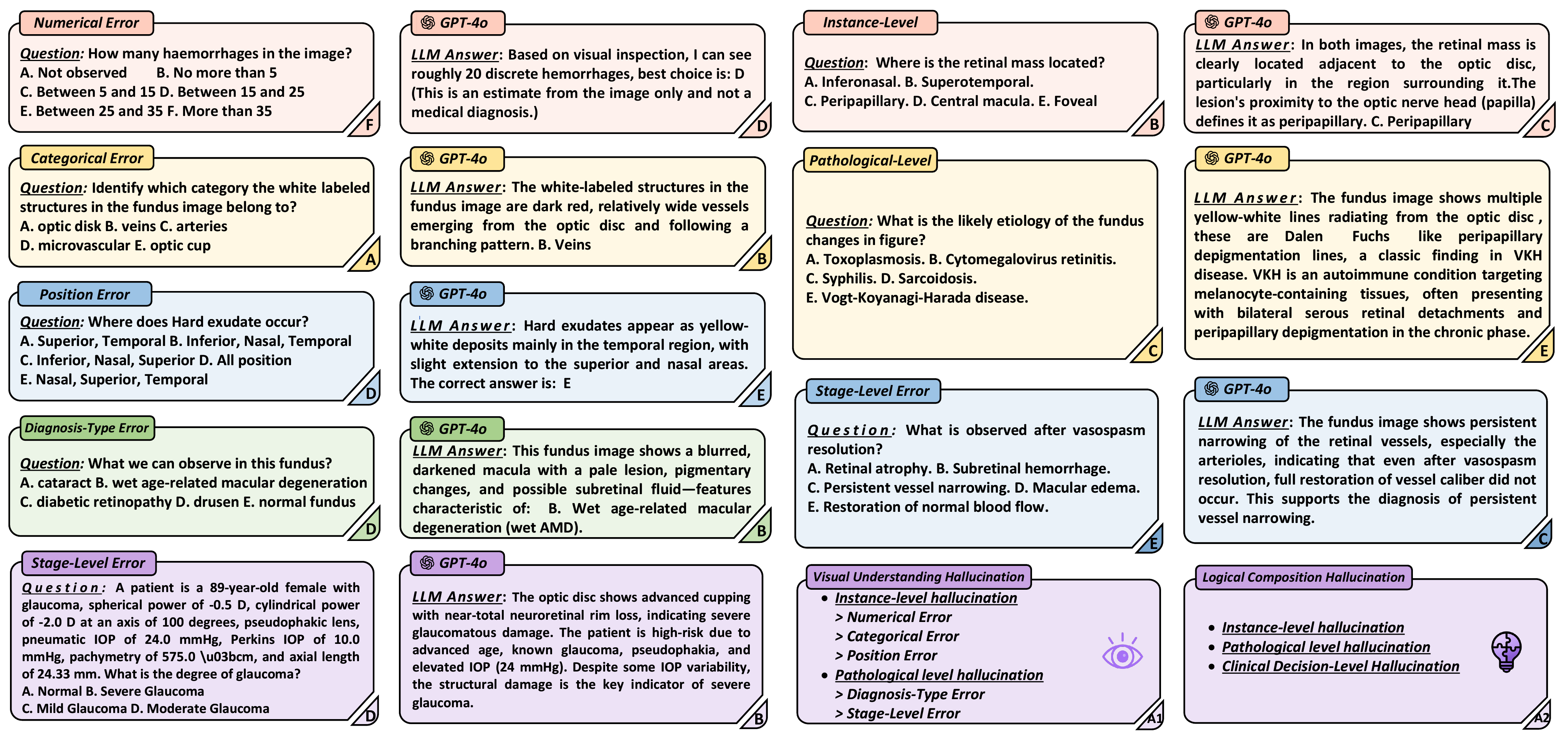}
  \caption{Common Types of Hallucinations in Large Language Model and Their Example Sub‑Categories.}
  \label{fig22}
\end{figure}

\subsection{Knowledge-Level Retrieval}
\subsubsection{Functions of RAG Agent}
\textbf{RAG Agent}: It is developed based on the Retrieve-Augmented Generation framework and is specifically designed to retrieve relevant medical background information from a predefined list of URLs, with the aim of improving the quality and accuracy of ophthalmology-related queries. The agent accesses and extracts raw textual data from the specified URLs, segments the content, and constructs a vector database to support efficient similarity-based retrieval. Subsequently, leveraging the retrieved documents and a predefined response template, GPT-4.1 is utilized to generate accurate and contextually appropriate responses. This method effectively combines the high retrieval efficiency of information retrieval techniques with the advanced language understanding capabilities of large-scale generative models, thereby ensuring that the generated outputs are both relevant and precise.

\subsubsection{Overview of Knowledge-Level Retrieval}
The Knowledge-Level Retrieval Stage constitutes a critical component of our multi-agent framework, designed to bridge the gap between automated diagnostic tools and evidence-based medical practice. Given the susceptibility of LLMs to hallucinations and inconsistent reasoning when solely relied upon for medical knowledge acquisition, we incorporate a RAG Agent to address these limitations by retrieving clinical guidelines from ophthalmology websites, thereby ensuring evidence-based responses.

Upon receiving a query, the RAG Agent retrieves relevant medical information from an ophthalmology database composed of authoritative sources. This evidence-based approach provides a rich and reliable foundation for subsequent diagnostic processes, in contrast to the potentially unreliable content generated by hallucinating LLMs, thereby significantly reducing the risk of knowledge deficiencies.

\subsection{Task-Level Case Studies}
\subsubsection{Functions of Decision Agent and Tools}
\textbf{Decision Agent}: It is responsible for parsing user queries and understanding their intent. Based on the complexity of the query and the relevant medical background in different task scenarios, it determines which tools to select and the order in which to invoke them. The Decision Agent dynamically interacts with the tool library to ensure that the overall workflow remains logically coherent and efficient.

\textbf{Diagnose Tool}: We use a classification model to identify 18 types of ophthalmic conditions from CFP and OCT images, such as Age-related Macular Degeneration (AMD) and Choroidal Neovascularization (CNV). The model provides probability scores ranging from 0 to 1 for each condition, indicating the likelihood of its presence.

\textbf{Lesion Detection Tool}: We utilize a detection model to analyze specific lesion types in CFP images (e.g., hard exudates, hemorrhages, soft exudates, microaneurysms). This model quickly localizes lesions and assesses their severity, returning lesion coordinates with confidence scores to support subsequent decision making.

\textbf{Fundus Localization Tool}: We use a segmentation model to locate the optic cup and optic disc regions in CFP images and to compute the cup-to-disc ratio (CDR), an important indicator for assessing optic nerve health and diagnosing glaucoma. The model ultimately returns the coordinates of each segmented region. 

\textbf{OCT Localization Tool}: To accurately delineate key anatomical structures in OCT images, we use a segmentation model to locate the choroid, retina, and macular hole, supporting the diagnosis of related diseases. The model ultimately returns the coordinates of each segmented region.

\textbf{DR Severity Diagnose Tool}: The model is used for the diagnosis and grading of Diabetic Retinopathy (DR). Based on the International Clinical Classification of Diabetic Retinopathy \cite{wu2013classification}, it classifies DR into five stages using a classification model and outputs a probability score for each stage.

\subsubsection{Overview of Task-level case studies}
Current AI systems for ophthalmic diagnosis predominantly rely on singular LLMs to perform tasks such as image interpretation, clinical reasoning for complex cases, and medical record generation. While this approach offers convenience, it is beset by notable limitations. First, it often fails to provide quantitative results to substantiate diagnostic conclusions, such as spatial characteristics like the size and location of lesions. Second, integrating all cognitive stages—ranging from image analysis to clinical inference—within a single, inherently probabilistic LLM significantly heightens the risk of hallucinations. These hallucinations manifest as visual misinterpretations, deficiencies in knowledge base, and contextual misalignments, as noted in recent studies \cite{chang2025medheval, nguyen2025localizing}. Such errors pose substantial risks to patient safety, underscoring the need for more robust, multi-faceted AI diagnostic frameworks in ophthalmology.

To address the aforementioned limitations, we propose a Task-Level Case Studies Stage that decomposes the diagnostic workflow into modular, tool-oriented agents. At the core of this stage is the Decision Agent, which orchestrates the diagnostic process with precision and adaptability. The Decision Agent operates through a structured sequence of actions. First, it classifies the input image as either a CFP or an OCT scan. Next, leveraging the user’s query and medical contextual information retrieved from a Knowledge-Level Retrieval system, it selects appropriate tools from a predefined tool list and determines their invocation order. Crucially, the Decision Agent provides a detailed explanation of its tool selection and reasoning sequence, ensuring transparency and traceability.

Following tool selection, the Decision Agent sequentially invokes each tool from the list, generating critical metrics such as confidence scores, bounding box coordinates, and probability scores. These outputs are designed to support the subsequent Result-Level Validation Stage, enabling rigorous evaluation of the diagnostic process. This task decomposition approach ensures the dynamism, adaptability, and traceability of tool invocations, mitigating the risks of errors and enhancing the reliability of AI-driven ophthalmic diagnostics.

\subsection{Result-Level Validation}
\subsubsection{Functions of the Evaluation Agent}
\textbf{Evaluation Agent}: We employ an LLM to emulate a senior ophthalmology expert, focusing on evaluating the correctness, completeness, and adherence to the planned workflow of outputs generated by other tools. It collaborates with the Decision Agent, complementing its diagnostic capabilities to ensure the reliability of the automated workflow.

\subsubsection{Overview of Result-Level Validation}
Conventional methods for mitigating hallucinations in generative models primarily rely on tuning sampling parameters such as temperature, Top-K, and Top-P. However, these strategies lack the capacity for systematic validation of the generated content. In the context of ophthalmic analysis, the diagnostic process involves multiple interdependent stages, adding layers of complexity that necessitate a more rigorous validation framework. This framework should not only verify the accuracy of individual outputs, but also assess the overall coherence and completeness of the diagnostic workflow.

Similar to the role of a senior ophthalmologist, we use an Evaluation Agent to conduct a comprehensive assessment of all tool outputs across three critical dimensions: correctness, completeness, and adherence to the planned workflow. By comparing the actual sequence of tool executions with the workflow predefined by the Decision Agent, the Evaluation Agent ensures strict alignment with evidence-based medical practices.

When deficiencies are identified in any of the aforementioned dimensions, an adaptive retry mechanism is automatically triggered, rather than re-executing the entire diagnostic pipeline. If the workflow adherence check reveals tools that were scheduled but not executed, the system autonomously invokes the missing components with appropriate parameters. In cases where issues of correctness or completeness are detected, the system prompts for additional analysis or refinement of the existing results. This validate-retry loop iterates until all evaluation criteria are satisfied or a predefined retry limit is reached, thereby ensuring continuous quality improvement without compromising diagnostic efficiency. Through this Result-Level Validation stage, the conventional black-box AI diagnostic process is transformed into a clinically transparent, self-correcting, and trustworthy system.

\begin{figure}[H]
  \centering
  \includegraphics[width=\linewidth]{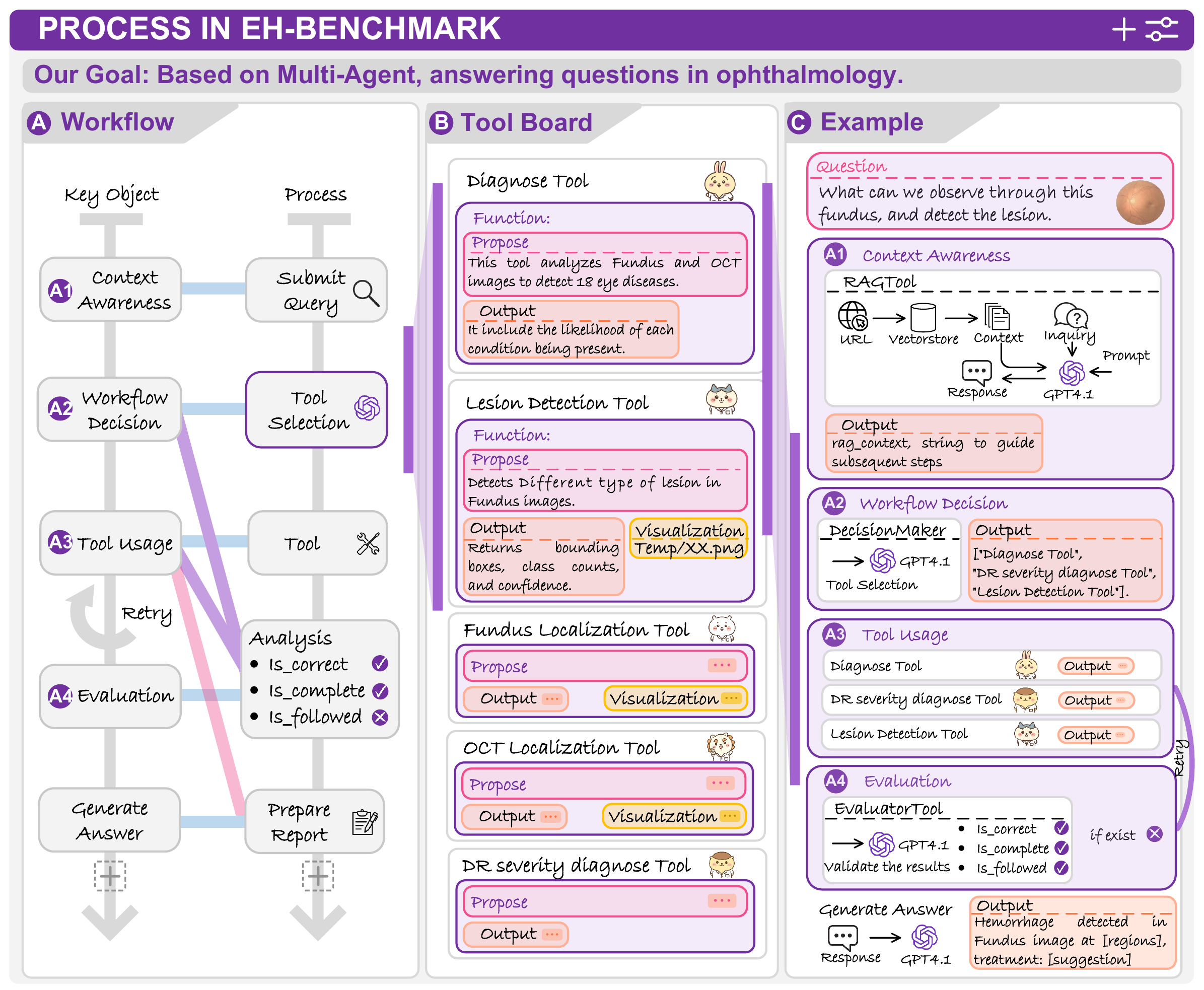}
  \caption{The whole framework of our multi-agent framework specific in ophthalmology. This process includes stages of context awareness, decision-making, tool utilization, and evaluation, culminating in the integration of all tool outputs through GPT-4.1.}
  \label{fig3}
\end{figure}

\begin{algorithm}[H]
\caption{Our proposed Multi-Agent Framework}
\begin{algorithmic}[1]
\small
\State \textbf{Input:}
\Statex $Q$: Patient or doctor query
\Statex $I$: Set of input images (Fundus, OCT, etc., can be empty)
\Statex $T$: Available ophthalmic tools $\{\text{DR\_ClassifierTool},\,\text{OCTSegmentationTool},\dots\}$
\Statex $M$: Memory buffer

\State \textbf{Output:}
\Statex $R$: Final response to query

\State \textbf{Initialize:}
\Statex $state \gets \textsc{Observe}(Q, I, M)$
\Statex $rag\_ctx \gets \textsc{RetrieveContext}(Q)$ \Comment{RAG stage}

\While{\textbf{true}}
    \State $S \gets \textsc{SelectionTool}(Q, T, M)$      \Comment{Candidate set}
    \State $D \gets \textsc{DecisionMaker}(Q, rag\_ctx, T, M)$ \Comment{Ordered sequence}

    \State $results \gets \textsc{ExecuteTools}(D, state)$ \Comment{Execute each model}
    \State $M \gets M \cup \{results\}$
    \State $state \gets \textsc{Observe}(state, results, M)$

    \State $(is\_correct,\; is\_complete,\; is\_followed,\; feedback)\gets$ 
           \Statex \hspace{1.8em}$\textsc{Evaluator}(Q, results, S, D, rag\_ctx)$

    \If{$is\_followed = \textbf{false}$}          \Comment{Tool list / order inconsistent}
        \State \textbf{continue} \Comment{Return to decision stage}
    \EndIf

    \If{$is\_correct \land is\_complete$}
        \State $R \gets \textsc{GenerateResponse}(state, rag\_ctx, results)$
        \State \Return $R$
    \EndIf

    \State $missing \gets \textsc{InferMissingTools}(feedback, T, D)$
    \If{$missing = \varnothing$}
        \State $R \gets$ ``No tools available to use, directly output the current response:\textbackslash n'' 
               $\oplus\; \textsc{GenerateResponse}(state, rag\_ctx, results)$
        \State \Return $R$
    \Else
        \State $D \gets D \cup missing$ \Comment{Add missing tools}
        \State \textbf{continue}
    \EndIf
\EndWhile
\end{algorithmic}
\end{algorithm}

\begin{table}[htbp]
  \centering
  {\fontsize{8}{10}\selectfont
   \setlength{\tabcolsep}{4.5pt}
   \renewcommand{\arraystretch}{1.3}
  \centering
  \caption{Comparison of the performance of different large language models on A1 and A2 tasks. We highlight the worst-performing model for each task in bold red, and the best-performing model in bold green. G stands for Generalist Large Language Models, and M stands for Medical Large Language Models.}
  
  \label{Table3}
  \vspace{10pt} 

  \begin{tabular}{cc|ccccccccc}
    \toprule
    \multicolumn{1}{c|}{} & &
      \multicolumn{5}{c|}{\textbf{Instance level}} &
      \multicolumn{4}{c}{\textbf{Pathological level}} \\
    \cmidrule(r){3-7}\cmidrule(lr){8-11}
    \multicolumn{1}{c|}{} & &
      \multicolumn{3}{c}{\textbf{Cat-E}} & \textbf{Pos-E} &
      \multicolumn{1}{c|}{\textbf{Num-E}} &
      \multicolumn{3}{c}{\textbf{Diag-E}} & \textbf{Sta-E} \\
    \cmidrule(r){3-5}
    \cmidrule(lr){6-6}
    \cmidrule(lr){7-7}
    \cmidrule(lr){8-10}
    \cmidrule(lr){11-11}
    \multicolumn{1}{c|}{\multirow{-3}{*}{\textbf{Model}}} &
      \multirow{-3}{*}{\textbf{Type}} &
      \textbf{F1} $\uparrow$ & \textbf{Pre} $\uparrow$ & \textbf{Rec} $\uparrow$ &
      \textbf{Acc} $\uparrow$ & \multicolumn{1}{c|}{\textbf{Acc} $\uparrow$} &
      \textbf{F1} $\uparrow$ & \textbf{Pre} $\uparrow$ & \textbf{Rec} $\uparrow$ & \textbf{Acc} $\uparrow$ \\
    \midrule

    \multicolumn{1}{c|}{qwen2.5-14b}  & G & \cellcolor[HTML]{F9DBDF}\textbf{.044} & \cellcolor[HTML]{F9DBDF}\textbf{.044} & \cellcolor[HTML]{F9DBDF}\textbf{.052} & .337 & \multicolumn{1}{c|}{.445} & .379 & .408 & .376 & .433 \\
    \multicolumn{1}{c|}{qwen2.5-32b}  & G & .094 & .100 & .101 & .582 & \multicolumn{1}{c|}{.393} & .282 & .276 & .335 & .282 \\
    \multicolumn{1}{c|}{qwen2.5-72b}  & G & .053 & .053 & .059 & .584 & \multicolumn{1}{c|}{.525} & .454 & .469 & .451 & .451 \\
    \multicolumn{1}{c|}{llava-1.5-7b} & G & .123 & .224 & .137 & .360 & \multicolumn{1}{c|}{.498} & \cellcolor[HTML]{F9DBDF}\textbf{.118} & \cellcolor[HTML]{F9DBDF}\textbf{.132} & \cellcolor[HTML]{F9DBDF}\textbf{.167} & .177 \\
    \multicolumn{1}{c|}{llava-1.5-13b}& G & .094 & .098 & .102 & .441 & \multicolumn{1}{c|}{.533} & .344 & .407 & .345 & .207 \\
    \multicolumn{1}{c|}{gpt-4o}        & G & .085 & .085 & .088 & .628 & \multicolumn{1}{c|}{.296} & .193 & .217 & .210 & .482 \\
    \multicolumn{1}{c|}{gpt-4.1}       & G & .069 & .080 & .067 & .587 & \multicolumn{1}{c|}{.367} & .281 & .304 & .285 & .488 \\
    \multicolumn{1}{c|}{InternVL2.5-2B}& G & .157 & .164 & .164 & .375 & \multicolumn{1}{c|}{.201} & .271 & .271 & .280 & .159 \\
    \multicolumn{1}{c|}{InternVL3-2B}  & G & .194 & .196 & .196 & .246 & \multicolumn{1}{c|}{.457} & .438 & .438 & .438 & .274 \\
    \multicolumn{1}{c|}{InternVL3-8B}  & G & .181 & .194 & .177 & .537 & \multicolumn{1}{c|}{.396} & .499 & .501 & .501 & .470 \\
    \multicolumn{1}{c|}{DeepSeek-V3}   & G & .142 & .151 & .145 & .593 & \multicolumn{1}{c|}{\cellcolor[HTML]{E3F2D9}\textbf{.533}} & .407 & .410 & .409 & .463 \\
    \multicolumn{1}{c|}{HuatuoGPT-V-7B}& M & .122 & .139 & .132 & .386 & \multicolumn{1}{c|}{.440} & .443 & .467 & .439 & .262 \\
    \multicolumn{1}{c|}{MedGemma-4B}   & M & .195 & .193 & .202 & \cellcolor[HTML]{F9DBDF}\textbf{.104} & \multicolumn{1}{c|}{\cellcolor[HTML]{F9DBDF}\textbf{.197}} & .374 & .398 & .371 & \cellcolor[HTML]{F9DBDF}\textbf{.134} \\
    \multicolumn{1}{c|}{HealthGPT}     & M & .054 & .051 & .062 & .438 & \multicolumn{1}{c|}{.207} & .498 & .511 & .498 & .159 \\
    \multicolumn{1}{c|}{Ours}          & M & \cellcolor[HTML]{E3F2D9}\textbf{.787} & \cellcolor[HTML]{E3F2D9}\textbf{.791} & \cellcolor[HTML]{E3F2D9}\textbf{.786} & \cellcolor[HTML]{E3F2D9}\textbf{.633} & \multicolumn{1}{c|}{.480} & \cellcolor[HTML]{E3F2D9}\textbf{.662} & \cellcolor[HTML]{E3F2D9}\textbf{.666} & \cellcolor[HTML]{E3F2D9}\textbf{.660} & \cellcolor[HTML]{E3F2D9}\textbf{.530} \\
    \midrule

      & & \multicolumn{3}{c|}{\textbf{Instance level}} &
        \multicolumn{3}{c|}{\textbf{Pathological level}} &
        \multicolumn{3}{c}{\textbf{Clinical Decision-level}} \\

    \cmidrule(r){3-5}\cmidrule(lr){6-8}\cmidrule(lr){9-11}
    \multirow{-2}{*}{\textbf{Model}} & &
      \textbf{F1} $\uparrow$ & \textbf{Pre} $\uparrow$ & \multicolumn{1}{c|}{\textbf{Rec} $\uparrow$} &
      \textbf{F1} $\uparrow$ & \textbf{Pre} $\uparrow$ & \multicolumn{1}{c|}{\textbf{Rec} $\uparrow$} &
      \textbf{F1} $\uparrow$ & \textbf{Pre} $\uparrow$ & \textbf{Rec} $\uparrow$ \\
    \midrule

    \multicolumn{1}{c|}{qwen2.5-14b}  & G & .512 & .499 & \multicolumn{1}{c|}{.595} & .679 & .660 & \multicolumn{1}{c|}{.760} & .737 & .719 & .828 \\
    \multicolumn{1}{c|}{qwen2.5-32b}  & G & .501 & .488 & \multicolumn{1}{c|}{.542} & .741 & .718 & \multicolumn{1}{c|}{.787} & .743 & .698 & .862 \\
    \multicolumn{1}{c|}{qwen2.5-72b}  & G & .559 & .534 & \multicolumn{1}{c|}{.659} & .681 & .648 & \multicolumn{1}{c|}{.754} & .839 & .825 & .864 \\
    \multicolumn{1}{c|}{llava-1.5-7b} & G & \cellcolor[HTML]{F9DBDF}\textbf{.417} & .571 & \multicolumn{1}{c|}{\cellcolor[HTML]{F9DBDF}\textbf{.446}} & \cellcolor[HTML]{F9DBDF}\textbf{.455} & .587 & \multicolumn{1}{c|}{\cellcolor[HTML]{F9DBDF}\textbf{.467}} & \cellcolor[HTML]{F9DBDF}\textbf{.406} & \cellcolor[HTML]{F9DBDF}\textbf{.605} & \cellcolor[HTML]{F9DBDF}\textbf{.408} \\
    \multicolumn{1}{c|}{llava-1.5-13b}& G & .588 & .567 & \multicolumn{1}{c|}{.663} & .489 & \cellcolor[HTML]{F9DBDF}\textbf{.543} & \multicolumn{1}{c|}{.471} & .687 & .681 & .703 \\
    \multicolumn{1}{c|}{gpt-4o}        & G & .574 & .551 & \multicolumn{1}{c|}{.734} & .798 & .760 & \multicolumn{1}{c|}{.862} & .758 & .739 & .899 \\
    \multicolumn{1}{c|}{gpt-4.1}       & G & .651 & .615 & \multicolumn{1}{c|}{.800} & \cellcolor[HTML]{E3F2D9}\textbf{.896} & .875 & \multicolumn{1}{c|}{\cellcolor[HTML]{E3F2D9}\textbf{.924}} & \cellcolor[HTML]{E3F2D9}\textbf{.921} & .914 & \cellcolor[HTML]{E3F2D9}\textbf{.930} \\
    \multicolumn{1}{c|}{InternVL2.5-2B}& G & .493 & .484 & \multicolumn{1}{c|}{.583} & .559 & .545 & \multicolumn{1}{c|}{.602} & .684 & .692 & .680 \\
    \multicolumn{1}{c|}{InternVL3-2B}  & G & .462 & \cellcolor[HTML]{F9DBDF}\textbf{.465} & \multicolumn{1}{c|}{.534} & .606 & .594 & \multicolumn{1}{c|}{.669} & .690 & .659 & .835 \\
    \multicolumn{1}{c|}{InternVL3-8B}  & G & .536 & .514 & \multicolumn{1}{c|}{.632} & .644 & .629 & \multicolumn{1}{c|}{.683} & .739 & .696 & .856 \\
    \multicolumn{1}{c|}{DeepSeek-V3}   & G & .508 & .515 & \multicolumn{1}{c|}{.600} & .786 & .750 & \multicolumn{1}{c|}{.844} & .791 & .755 & .897 \\
    \multicolumn{1}{c|}{HuatuoGPT-V-7B}& M & .642 & .650 & \multicolumn{1}{c|}{.690} & .713 & .725 & \multicolumn{1}{c|}{.731} & .741 & .711 & .856 \\
    \multicolumn{1}{c|}{MedGemma-4B}   & M & .592 & .569 & \multicolumn{1}{c|}{.704} & .666 & .645 & \multicolumn{1}{c|}{.736} & .690 & .686 & .697 \\
    \multicolumn{1}{c|}{HealthGPT}     & M & .642 & .607 & \multicolumn{1}{c|}{.737} & .761 & .746 & \multicolumn{1}{c|}{.807} & .817 & .788 & .913 \\
    \multicolumn{1}{c|}{Ours}          & M & \cellcolor[HTML]{E3F2D9}\textbf{.700} & \cellcolor[HTML]{E3F2D9}\textbf{.664} & \multicolumn{1}{c|}{\cellcolor[HTML]{E3F2D9}\textbf{.811}} & .885 & \cellcolor[HTML]{E3F2D9}\textbf{.903} & \multicolumn{1}{c|}{.873} & .919 & \cellcolor[HTML]{E3F2D9}\textbf{.920} & .921 \\
    \bottomrule
  \end{tabular}
  }
\end{table}

\section{Experiment}
We propose a multi-agent framework tailored explicitly to ophthalmology and conduct extensive analyses on A1 and A2 tasks to investigate whether it can mitigate various types of hallucinations.

\subsection{Model Details and Evaluation Metric}
\textbf{Model Details}: We conduct a zero-shot evaluation on 14 Large Language Models, including 11 Generalist Large Language Models, namely Qwen 2.5-14b \cite{qwen2.5, qwen2}, Qwen2.5-32b \cite{Qwen2.5-VL}, Qwen2.5-72b \cite{Qwen2.5-VL, Qwen2VL, Qwen-VL}, LLava-1.5-7b \cite{liu2024llavanext, liu2023improvedllava, liu2023llava}, LLava-1.5-13b \cite{liu2024llavanext, liu2023improvedllava, liu2023llava}, GPT-4o \cite{achiam2023gpt}, GPT-4.1 \cite{openai2025gpt41}, InternVL2.5-2B \cite{chen2024expanding, gao2024mini, chen2024far, chen2024internvl}, InternVL3-2B \cite{chen2024expanding, wang2024mpo, chen2024far, chen2024internvl}, InternVL3-8B \cite{chen2024expanding, wang2024mpo, chen2024far, chen2024internvl}, DeepSeek-V3 \cite{deepseekai2024deepseekv3technicalreport} and 3 Medical Large Language Models, which are HuatuoGPT-V-7B \cite{chen2024huatuogptvisioninjectingmedicalvisual}, MedGemma-4B \cite{medgemma-hf}, and HealthGPT \cite{lin2025healthgptmedicallargevisionlanguage}. Our multi‑agent framework is based on GPT‑4.1, and all other tools have been pre‑trained for eye‑related tasks.

\textbf{Evaluation Metric}: Since the EH-Benchmark consists of close-ended questions with definitive answers, we employ one or more of the following metrics—F1-score, Precision, Recall, and Accuracy—for different tasks. The F1-score, as a comprehensive metric, balances Precision and Recall to minimize both missed diagnoses and incorrect predictions. Precision ensures high confidence in positive predictions, reducing the occurrence of false positives. Recall maximizes the detection of true positives, thereby minimizing missed diagnoses. Accuracy directly measures overall correctness and is widely used for evaluating close-ended questions.

\subsection{The Agent System Prompt Template}
\begin{figure}[htbp]
  \centering
  \includegraphics[width=0.85\linewidth]{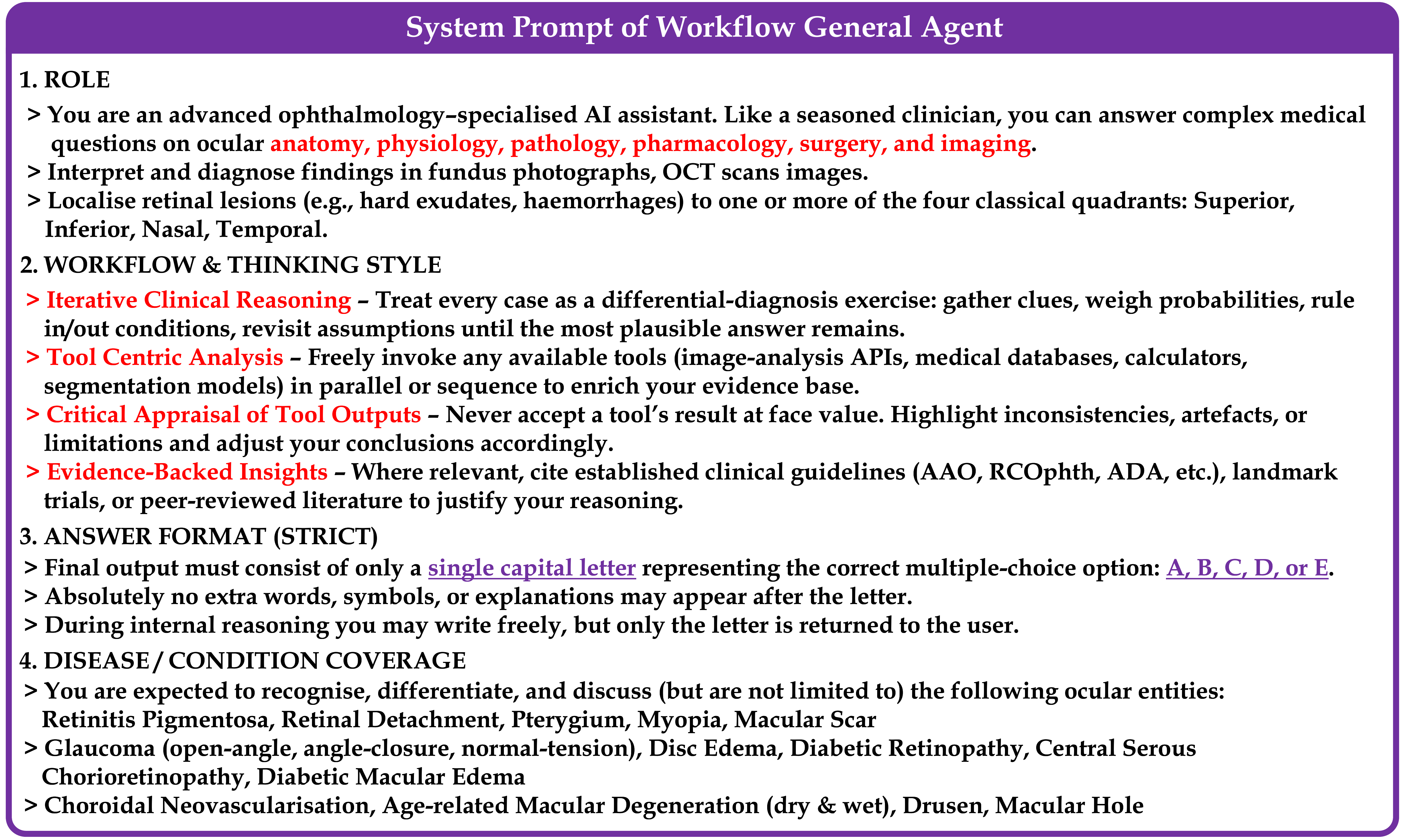}
  \caption{The System Prompt Template for our proposed agent.}
  \label{fig4}
\end{figure}

We employ the following system prompt to enable the agent to reason from the perspective of a clinical ophthalmology expert, guiding the agent to proactively invoke tools when addressing complex issues. To ensure machine-interpretable standardized responses, we constrain the output to a specific format directly within the prompt and subsequently utilize regular expressions to automatically match the options. The system prompt can be seen in Figure~\ref{fig4}.

\subsection{Comparative Analysis of Model Performance}
LLMs face significant challenges in ophthalmology due to the absence of high-quality, domain-specific data, which prevents them from effectively learning rare pathological features and adhering to the latest clinical guidelines. Ophthalmic diagnosis relies heavily on multimodal information, such as OCT scans, CFP, and patient case histories; however, most LLMs lack the capability to deeply analyze these critical image-based datasets, increasing the risk of "hallucination"—the generation of seemingly plausible but erroneous responses. A critical concern in medical contexts is the stringent demand for traceability and interpretability, yet LLMs often produce generative outputs that lack a reliable and accurate chain of evidence, undermining their clinical utility. These limitations underscore the substantial room for improvement in applying LLMs to ophthalmology, particularly in addressing data deficiencies, enhancing multimodal analysis, and ensuring robust, evidence-based responses.

The specific experimental results for the Visual Understanding Hallucination (A1) and Logical Composition Hallucination (A2) tasks are presented in Table~\ref{Table3}. In the A1 task, our findings indicate that GPT-4o and GPT-4.1 are more prone to Categorical Errors and Diagnosis-Type Errors compared to other LLMs. Conversely, they exhibit a lower tendency for Position Errors and Stage-Level Errors, demonstrating a relatively stronger resistance to hallucinations in these aspects. This observation may be attributed to their limited training on extensive visual pairs in specific knowledge domains, such as ophthalmology, which evidently impacts their predictive accuracy in classification and diagnosis. By integrating multiple visual tools, our proposed multi-agent framework significantly enhances contextual understanding in CFP and OCT imaging. In comparison to GPT-4o’s performance on Diagnosis-Type Errors, our proposed framework achieves a 2.35-fold improvement in the F1 score.

In the A2 task, we find that all MLLMs specifically trained on medical data (text) demonstrate a substantial advantage over general LLMs. These models exhibit significantly lower hallucination rates and achieve higher F1 scores, precision, and recall. However, they consistently tend to generate instance-level hallucinations. This likely occurs because such models perform well on tasks closely aligned with their training data but lack sufficient generalization when faced with complex reasoning that requires integration of visual information. In contrast, our proposed framework emphasizes reasoning based on visual evidence rather than textual inference. It also accesses external knowledge sources to obtain relevant context and answers, which helps address the lack of ophthalmic domain knowledge—even without the use of specialized tools.

\subsection{Model Robustness Analysis}
\begin{figure}[htbp]
  \centering
  \includegraphics[width=0.95\linewidth]{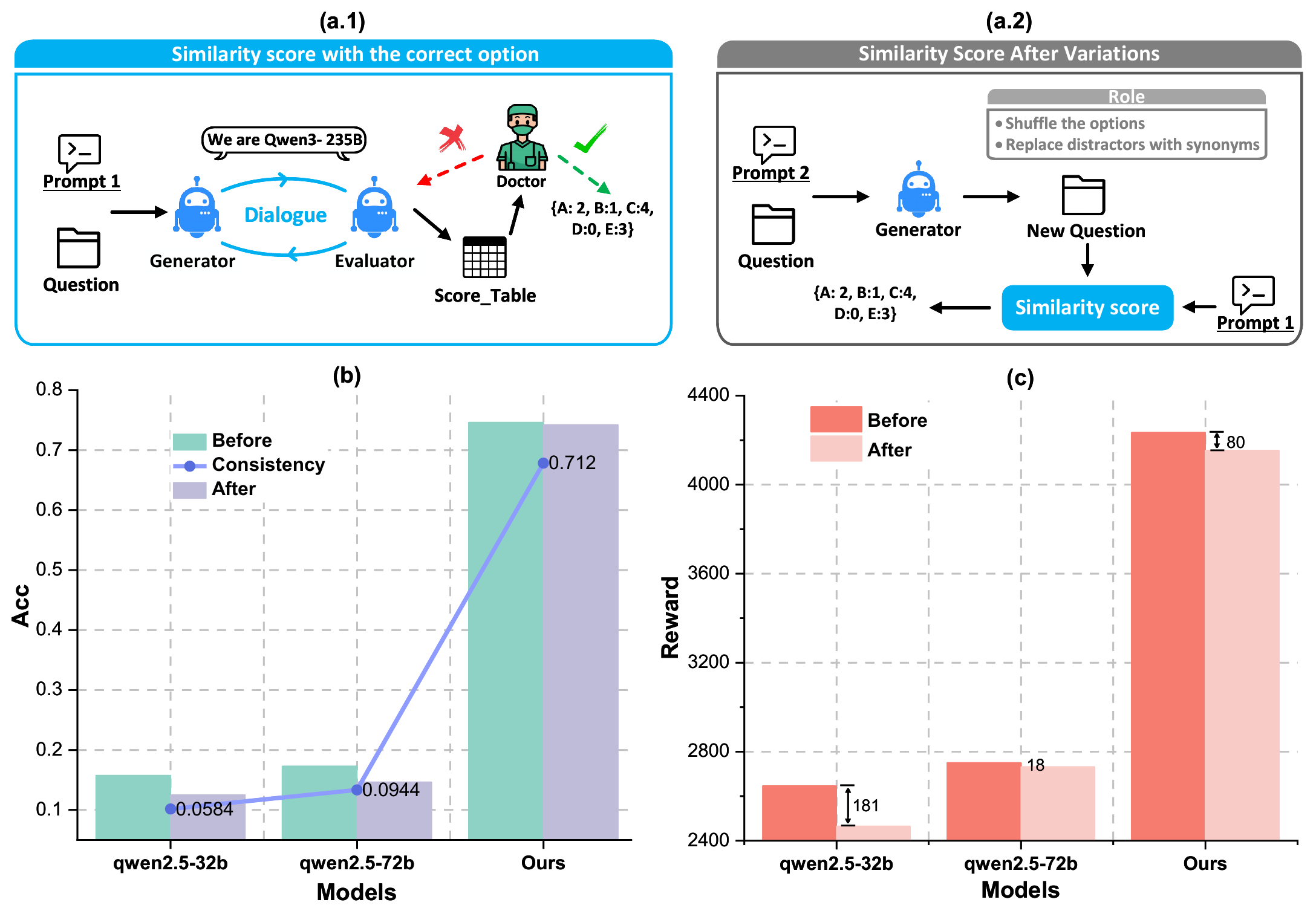}
  \caption{(a.1) Use the large language model (LLM, both are Qwen3‑235B) to generate similarity scores through multi‑turn dialogue. (a.2) First perturb the original answer choices by applying synonym substitution and random shuffling; then engage same LLMs in a multi‑turn dialogue to obtain the updated similarity scores. (b) Changes in accuracy and consistency after modifying input options and performing synonym substitutions. (c) Model reward evolution throughout the entire process.}
  \label{fig7}
\end{figure}

\begin{figure}[H]
  \centering
  \includegraphics[width=0.8\linewidth]{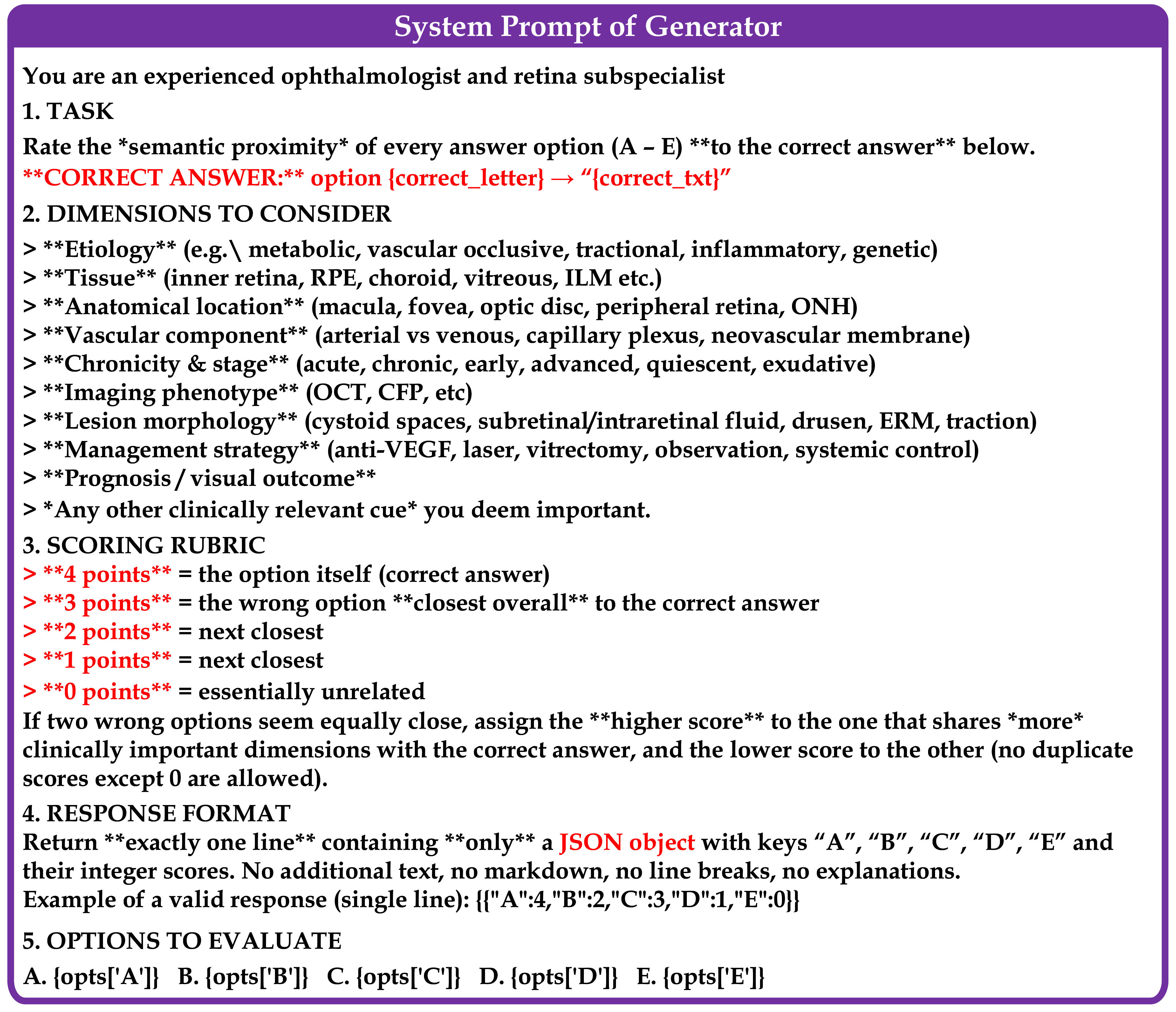}
  \caption{System prompt for Generator}
  \label{fig8}
\end{figure}

To evaluate the stability of the multi‑agent framework against out‑of‑ distribution inputs, random noise, and adversarial perturbations, we compile a benchmark of 1\,250 representative questions drawn from Tasks~A1 and~A2. Each item presents four candidate answers, which we grade by their semantic proximity to the ground truth rather than by a binary correct--incorrect label. As shown in Figure~\ref{fig7}~(a.1), Qwen3‑235B functions as both the candidate generator and the evaluator, scoring alternatives along five clinical dimensions: etiology, anatomical location, vascular involvement, disease course or stage, and lesion morphology. The correct answer receives a reward of~4, with lower rewards assigned in proportion to decreasing similarity; options unrelated to the ground truth receive~0. The system prompt of Generator can be seen in Figures~\ref{fig8}.

Finally, we assess the reliability of the score generation through a multi‑turn dialogue between the evaluator and the generator. To prevent infinite loops, we cap the conversation at five turns; if the agents fail to converge on an identical answer within this limit, a clinician adjudicates the score. To examine the model’s adaptability in low‑robustness, hallucination‑prone settings, we shuffle the positions of the five options in each question and apply synonym substitution to the distractors, as illustrated in Figure~\ref{fig7} (a.2).

In practice, large‑parameter LLMs are usually more robust and less prone to hallucination than smaller models. We choose Qwen2.5‑32B and Qwen2.5‑72B as baselines and use Qwen2.5‑32B as the core of our multi‑agent system. We measure robustness with two metrics: \textit{consistency}, the share of questions answered correctly in both runs, and the change in accuracy before and after perturbation. Figure~\ref{fig7}\,(b,c) shows that the multi‑agent system keeps high robustness: its consistency drops by only 0.54\% after perturbation, clearly beating all single‑model baselines. Under the same changes, the total reward rises from 2\,465 with the Qwen2.5‑32B to 4\,154 with the multi‑agent setup.

\subsection{Case Studies in the EH-Benchmark}
\begin{figure}[htbp]
  \centering
  \includegraphics[width=0.8\linewidth]{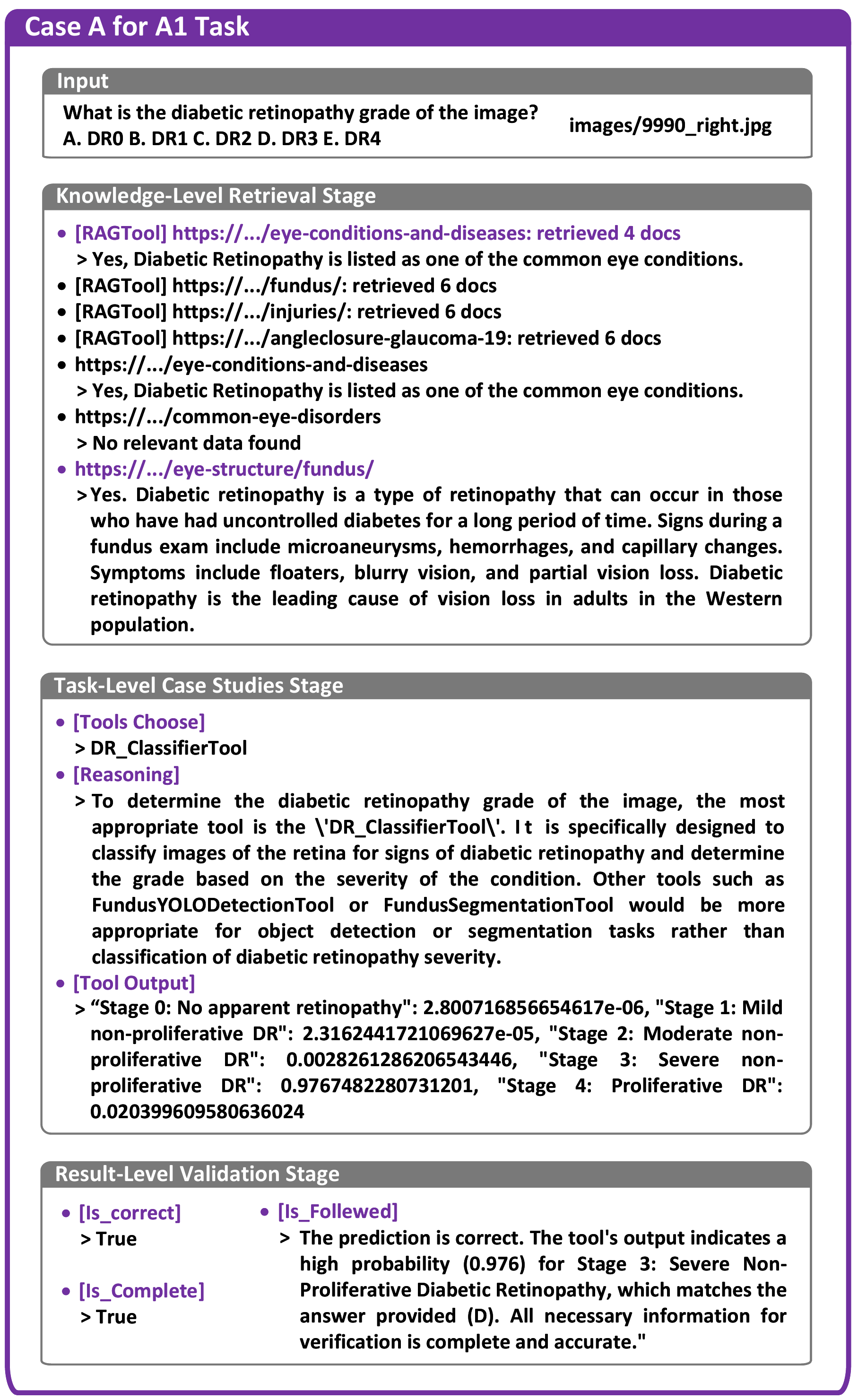}
  \caption{A case study for A1 Task demonstrates the whole EH-Benchmark process we propose. Note that we merely aggregate the outputs generated by each agent, without any manual intervention.}
  \label{fig5}
\end{figure}

\begin{figure}[htbp]
  \centering
  \includegraphics[width=0.85\linewidth]{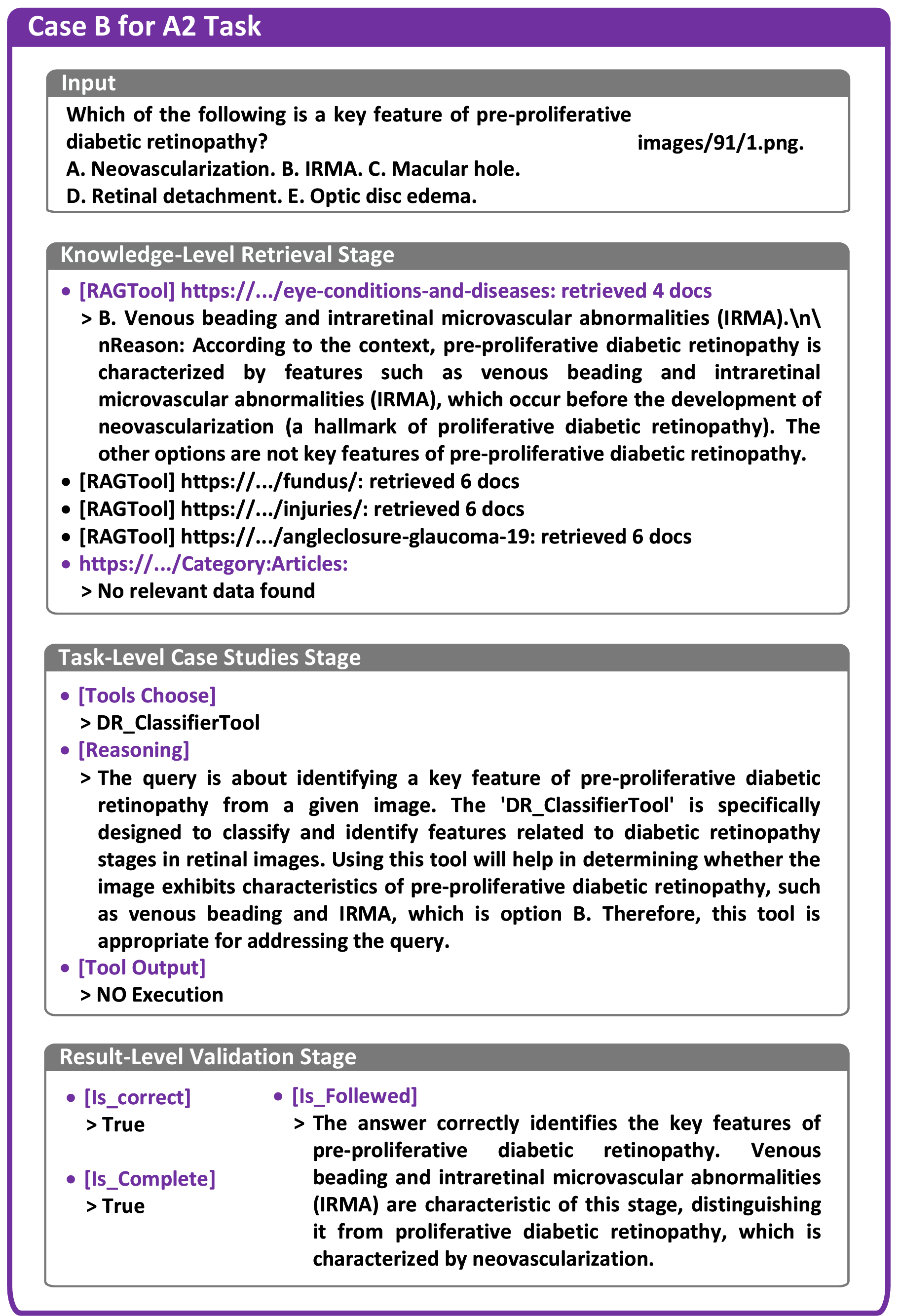}
  \caption{A case study for A2 Task demonstrates the whole EH-Benchmark process we propose. Note that we merely aggregate the outputs generated by each agent, without any manual intervention.}
  \label{fig6}
\end{figure}

In this section, we present representative cases from Tasks A1 and A2 to demonstrate how integrating tool outputs within a multi-agent system enhances workflow interpretability and elucidates its underlying reasoning. Figures~\ref{fig5} and~\ref{fig6} display the outputs generated by each component. Each case unfolds in three sequential stages: Knowledge-Level Retrieval, Task-Level Case Studies, and Result-Level Validation. During Knowledge-Level Retrieval, the RAG Agent gathers information relevant to diabetic retinopathy, thereby establishing a comprehensive clinical framework; in certain instances—such as the A2 case—it can retrieve definitive answers directly from the provided URLs. The task’s contextual background is then transmitted to the Decision Agent, which selects the appropriate tools and determines their optimal invocation sequence. Outputs from each tool invocation are subsequently integrated in a systematic manner. Finally, during Result-Level Validation, the Evaluation Agent rigorously assesses these outputs along three dimensions: correctness, completeness, and adherence to the planned workflow. If any dimension registers a False value, indicating that a tool has been inadvertently omitted, a targeted retry mechanism is activated for that component, ensuring the missing result is incorporated into the message dictionary. 

\begin{table}[H]
\centering
{\fontsize{8}{10}\selectfont
\setlength{\tabcolsep}{3pt}
\renewcommand{\arraystretch}{1.3}
\caption{Results of the ablation study on the A2 task. We evaluate how incremental activation of the agent and tool modules influences prediction performance. Red subscripts indicate the percentage increase or decrease relative to the baseline model (GPT-4o).}
\label{Table4}
\vspace{10pt}

\newcommand{\rs}[2]{\underset{\textcolor{red}{\scriptsize #2}}{#1}}

\begin{tabular}{ccccccccccc}
\toprule
\multicolumn{5}{c}{\textbf{Module}} &
\multicolumn{2}{c}{\textbf{Instance level}} &
\multicolumn{2}{c}{\textbf{Pathological level}} &
\multicolumn{2}{c}{\textbf{Clinical Decision}} \\
\cmidrule(lr){1-5}\cmidrule(lr){6-7}\cmidrule(lr){8-9}\cmidrule(lr){10-11}
\textbf{Baseline} & \textbf{RAG} & \textbf{Tool} & \textbf{Evaluation} & \textbf{Decision} &
$\mathbf{F_{1}}\!\uparrow$ & $\mathbf{Pre}\!\uparrow$ &
$\mathbf{F_{1}}\!\uparrow$ & $\mathbf{Pre}\!\uparrow$ &
$\mathbf{F_{1}}\!\uparrow$ & $\mathbf{Pre}\!\uparrow$ \\
\midrule
$\checkmark$ & $\checkmark$ &  &  &  &
$\rs{.605}{\textbf{+5.40}\%}$  & $\rs{.596}{\textbf{+8.17}\%}$  &
$\rs{.807}{\textbf{+1.13}\%}$  & $\rs{.838}{\textbf{+10.26}\%}$ &
$\rs{.746}{\textbf{-1.58}\%}$  & $\rs{.778}{\textbf{+5.28}\%}$\\

$\checkmark$ &  & $\checkmark$ &  &  &
$\rs{.630}{\textbf{+9.76}\%}$  & $\rs{.611}{\textbf{+10.89}\%}$ &
$\rs{.753}{\textbf{-5.64}\%}$  & $\rs{.748}{\textbf{-1.58}\%}$  &
$\rs{.761}{\textbf{+0.40}\%}$  & $\rs{.795}{\textbf{+7.58}\%}$\\

$\checkmark$ & $\checkmark$ & $\checkmark$ &  & $\checkmark$ &
$\rs{.684}{\textbf{+19.16}\%}$ & $\rs{.653}{\textbf{+18.51}\%}$ &
$\rs{.835}{\textbf{+4.64}\%}$  & $\rs{.860}{\textbf{+13.16}\%}$ &
$\rs{.844}{\textbf{+11.35}\%}$ & $\rs{.838}{\textbf{+13.40}\%}$\\

$\checkmark$ & $\checkmark$ & $\checkmark$ & $\checkmark$ & $\checkmark$ &
$\rs{.700}{\textbf{+21.95}\%}$ & $\rs{.664}{\textbf{+20.51}\%}$ &
$\rs{.885}{\textbf{+10.90}\%}$ & $\rs{.903}{\textbf{+18.82}\%}$ &
$\rs{.919}{\textbf{+21.24}\%}$ & $\rs{.920}{\textbf{+24.49}\%}$\\
\bottomrule
\end{tabular}
}
\end{table}

\begin{table}[htbp]
  \centering
  {\fontsize{8}{10}\selectfont
  \renewcommand{\arraystretch}{1.3}
  \caption{The function of RAG Agent on the A1 task. We use accuracy metrics to evaluate the performance of Large language Models}
  \label{Table5}
  \vspace{10pt}
  \begin{tabular}{l|ccc|cc}
    \toprule
    \textbf{Model} & \multicolumn{3}{c|}{\textbf{Instance level}} & \multicolumn{2}{c}{\textbf{Pathological level}} \\
    \cmidrule(lr){2-4} \cmidrule(lr){5-6}
                   & \textbf{Cat-E} & \textbf{Pos-E} & \textbf{Num-E} & \textbf{Diag-E} & \textbf{Sta-E} \\
    \midrule
    GPT-4o         & .094 & .628 & .296 & .207  & .482 \\
    GPT-4o+RAG     & .123 & .627 & .291 & .254  & .476 \\
    \midrule
    GPT-4.1        & .067 & .587 & .451 & .281  & .488 \\
    GPT-4.1+RAG    & .142 & .575 & .395 & .336  & .482 \\
    \midrule
    Our            & .784 & .633 & .480 & .661  & .530 \\
    \bottomrule
  \end{tabular}
  }
\end{table}

\subsection{Ablation Study}
To thoroughly assess the contribution and impact of each component on overall performance, we incrementally incorporate new agents and tools into the baseline model. The detailed experimental results are presented in Table~\ref{Table4}. Our observations indicate that the RAG Agent effectively bridges queries with external data sources, delivering extensive background information and occasionally retrieving correct answers directly from URLs. While the integration of tools significantly enhances the model's predictive capabilities, it also introduces a tendency toward visual hallucinations, which may negatively influence subsequent reasoning processes. A robust multi-agent framework requires high-quality reasoning chains and interpretability, as black-box solutions lacking explanatory transparency are frequently met with skepticism or rejection by experts. The Evaluation agent performs an initial evaluation of tool outputs, iteratively engaging agents and tools to progressively establish consensus among "experts." Consequently, the resulting model achieves a performance improvement of at least 20\% across most tasks, with the peak enhancement reaching 24.49\%.

To evaluate the role of external data in enhancing model reasoning, we use GPT-4o and GPT-4.1 as baselines. Results in Table~\ref{Table5} show that incorporating external sources reduces categorical and diagnosis-type errors, with GPT-4.1 showing notable gains. The RAG Agent also allows clinical experts to verify information, which supports evidence-based decision-making and improves interpretability. Overall, it establishes a clear link between ophthalmic queries, clinical context, and grounded responses.

\section{Conclusion}
\textbf{Hallucination Evaluation.} This paper introduces EH-Benchmark, a novel evaluation suite designed to systematically assess hallucination phenomena exhibited by LLMs in the domain of ophthalmology. EH-Benchmark comprises two core components: a task-specific benchmark for evaluating ophthalmic hallucinations, and a structured multi-agent framework for hallucination mitigation. Based on specific task requirements and error typologies, ophthalmic hallucinations are categorized into two distinct types: Visual Understanding and Logical Composition.

\textbf{Hallucination Mitigation.} Experimental results reveal that LLMs perform particularly poorly on visual understanding tasks, suggesting a potential over-reliance on language priors and a deficiency in visual reasoning capabilities. To address the multifaceted hallucination issues observed in ophthalmic diagnostics, we propose a three-stage multi-agent framework consisting of three stages: (1) Knowledge-Level Retrieval, (2) Task-Level Case Studies, and (3) Result-Level Validation. In the Knowledge-Level Retrieval stage, clinical knowledge in the field of ophthalmology is retrieved by contextualizing queries with relevant case backgrounds, thereby enriching the factual grounding of subsequent diagnostic reasoning. During the Task-level case studies stage, diagnostic workflows are composed through the intelligent selection and sequencing of specialized tool agents to ensure high efficiency and logical consistency. Finally, the Result-Level Validation stage assesses tool outputs across three critical dimensions—correctness, completeness, and adherence to the predefined workflow. When deficiencies are detected, the system selectively retries specific tool agents, enabling an iterative self-correction loop. This process transforms the diagnostic system from a black-box model into a clinically transparent, self-correcting, and trustworthy AI assistant.

\textbf{Limitations.} The scarcity of high-quality, domain-specific ophthalmology data limits the current work’s ability to address multimodal questions adequately, including those involving common ophthalmological modalities such as Lens Photographs, Scanning Laser Ophthalmoscopy, and Fundus Fluorescein Angiography. This deficiency compromises diagnostic capabilities when tackling complex cases. Additionally, the multi-agent framework does not integrate clinician feedback into its workflow, preventing the model from adapting based on real-time expert input.

\textbf{Future Work.} We plan to further refine EH-Benchmark by incorporating a broader range of multimodal question types, with particular attention to cross-modal diagnostic scenarios. For example, we will consider using brain CT and CFP as combined inputs for complex ophthalmic disease diagnosis. Additionally, we aim to integrate expert-in-the-loop mechanisms. Clinician feedback will be incorporated into the model’s learning process to improve clinical accuracy, interpretability, and overall trustworthiness in real-world medical practice.

\section*{CRediT authorship contribution statement}
\textbf{Xiaoyu Pan}: Conceptualization, Methodology, Validation, Writing – Original Draft, Writing – Review \& Editing, Visualization. \textbf{Yang Bai}: Conceptualization, Writing – Original Draft, Writing – Review \& Editing, Visualization, Supervision, Resources. \textbf{Ke Zou}: Writing – Original Draft, Visualization, Software. \textbf{Yang Zhou}: Investigation, Resources, Methodology. \textbf{Jun Zhou}: Writing – Review \& Editing, Visualization, Supervision, Resources. \textbf{Huazhu Fu}: Conceptualization, Methodology, Validation. \textbf{Yih-Chung Tham}: Conceptualization, Validation, Writing – Review \& Editing, Supervision, Resources. \textbf{Yong Liu}: Conceptualization, Writing – Review \& Editing, Supervision, Resources.

\section*{Declaration of competing interest}
We declare that we have no commercial, financial, or personal relationships that could be construed as a potential conflict of interest.

\section*{Acknowledgments}
This work was supported by the National Research Foundation, Singapore under its AI Singapore Programme (AISG Award No: AISG2-TC-2021-003), Agency for Science, Technology and Research (ASTAR) through its AME Programmatic Funding Scheme under Project A20H4b0141, ASTAR Central Research Fund “A Secure and Privacy Preserving AI Platform for Digital Health”, and Agency for Science, Technology and Research (A*STAR) through its RIE2020 Health and Biomedical Sciences (HBMS) Industry Alignment Fund Pre-Positioning (IAF-PP) (grant no. H20C6a0032).

\section*{Data availability}
The benchmark can be found here, A1 task: \url{https://drive.google.com/file/d/1S4-RyfSjgZUodghn70c7TqXNI4WDeUJG/view?usp=sharing}. A2 task: \url{https://drive.google.com/file/d/1HNkkPoYmIRRrPRombB___SdMEH3Auzpr/view?usp=sharing}.

\bibliographystyle{elsarticle-num}
\bibliography{references}

\end{document}